\documentclass[10pt, conference, compsocconf]{IEEEtran}
\IEEEoverridecommandlockouts

\usepackage{cite}

\ifCLASSINFOpdf
  \usepackage[pdftex]{graphicx}
  \graphicspath{{./pics/}}
  \DeclareGraphicsExtensions{.pdf,.jpeg,.jpg,.png}
\else
\fi

\usepackage{xspace}

\usepackage[utf8]{inputenc}

\usepackage{amssymb}
\usepackage{amsmath, amsthm}
\usepackage[hidelinks, bookmarks=false]{hyperref} 
\usepackage{mathabx}

\usepackage{threeparttable}
\usepackage{makecell}
\usepackage{multirow}
\usepackage{hhline}

\newcommand*{\percent}[1]{$#1\,\%$}

\newcommand{\comment}[1]{\ignorespaces}

\usepackage{ifthen}
\newcommand{\blind}{\boolean{false}} 

\begin{document}
%
\title{Evaluating Sequence-to-Sequence Models for Handwritten Text Recognition}

\ifthenelse{\blind}
{\author{\textit{BLIND AUTHORS}}}
{\author{
	\IEEEauthorblockN{Johannes Michael, Roger Labahn}
	\IEEEauthorblockA{Computational Intelligence Technology Lab\\
		University of Rostock\\
		18057 Rostock, Germany\\
		\{johannes.michael,roger.labahn\}@uni-rostock.de
	}
	\and
	\IEEEauthorblockN{Tobias Grüning, Jochen Zöllner}
	\IEEEauthorblockA{PLANET artificial intelligence GmbH\\
		Warnowufer 60\\
		18057 Rostock, Germany\\
		\{tobias.gruening,jochen.zoellner\}@planet.de
	}
}}

\maketitle
\begin{abstract}
Encoder-decoder models have become an effective approach for sequence learning tasks like machine translation, image captioning and speech recognition, but have yet to show competitive results for handwritten text recognition. To this end, we propose an attention-based sequence-to-sequence model. It combines a convolutional neural network as a generic feature extractor with a recurrent neural network to encode both the visual information, as well as the temporal context between characters in the input image, and uses a separate recurrent neural network to decode the actual character sequence. We make experimental comparisons between various attention mechanisms and positional encodings, in order to find an appropriate alignment between the input and output sequence. The model can be trained end-to-end and the optional integration of a hybrid loss allows the encoder to retain an interpretable and usable output, \comment{e.g. for keyword spotting purposes without prior indexing,} if desired. We achieve competitive results on the IAM and ICFHR2016 READ data sets compared to the state-of-the-art without the use of a language model, and we significantly improve over any recent sequence-to-sequence approaches.
\end{abstract}

\renewcommand\IEEEkeywordsname{Keywords}
\begin{IEEEkeywords}
sequence-to-sequence, Seq2Seq, encoder-decoder, attention, handwritten text recognition, HTR
\end{IEEEkeywords}

\IEEEpeerreviewmaketitle

\section{Introduction}
Numerous historical collections from different time periods and locations were digitized at a great expense during the last decades. Millions of pages are scanned and available as images. To enable humanists, historians, genealogists as well as ordinary people to efficiently work with these documents, it is subject to current research and scientific discussion to make the content of these documents digitally available.

Since 2009, tremendous progress in the field of Handwritten Text Recognition (HTR) \comment{\cite{GravesSchmidhuber2009,LeifertSGWL16}} and Keyword Spotting (KWS) \comment{\cite{PuigcerverTV2014,StraussGLL2016,StraussLGL2016}} was achieved. Nowadays performance of deep learning based state-of-the-art systems reaches character error rates below \percent{10} for HTR \cite{SanchezRTV2014} and mean average precisions above $0.9$ for KWS \cite{PratikakisZPTV2016}. Nevertheless, due to inherent differences in the writing of individuals and the vague nature of handwritten characters, HTR remains a challenging open research problem, where robustness and adaptivity require further improvement.

Generally, the de facto standard for 
HTR tasks have been systems based on Convolutional Neural Networks (CNNs) and Recurrent Neural Networks (RNNs) \cite{Graves2009, Puigcerver2017}, which are utilizing the Connectionist Temporal Classification (CTC) \cite{GravesFGS2006} objective function. However, CTC-based architectures are subject to inherent limitations like strict monotonic input-output alignments and an output sequence length that is bound by the, possibly subsampled, input length. On the contrary, sequence-to-sequence (Seq2Seq) models that follow the encoder-decoder framework \cite{SutskeverVL2014} \comment{\cite{ChoMGBSB2014}} are more flexible, suit the temporal nature of text and are able to focus on the most relevant features of the input by incorporating attention mechanisms \cite{BahdanauCB2014}. Additionally, attention enables the networks to potentially model language structures, rather than simply mapping an input to an output \cite{ChoCB2015}.
Moreover, the latest advances in machine translation \cite{BahdanauCB2014, SutskeverVL2014}, image captioning \cite{XuBKCCSZB2015} and speech recognition \cite{BahdanauCSBB2015} \comment{\cite{ChorowskiBSCB2015}} motivate a systematic investigation of such models in the context of HTR tasks.

We propose a Seq2Seq model consisting of a deep CNN-RNN-encoder and an RNN-decoder that can be trained end-to-end on full line or word labels. The encoder extracts low-level features from the written text line and sequentially encodes temporal context between them. The decoder outputs a character sequence one step at a time, using an attention mechanism to focus on the most relevant encoded features at each decoding step. The model can be built and trained in a way, that preserves an interpretable and usable encoder output, e.g. for KWS purposes without prior indexing \cite{strauss2017}. Though, evaluating KWS performance for our proposed model is out of the scope of this paper. We show that our Seq2Seq model reaches competitive results on various HTR data sets, beating any Seq2Seq-based architectures that we know of \cite{PoulosValle2017, ChowdhuryVig2018, SueirasRSV2018, KangTRVFR2018}.

This is achieved by a thorough evaluation of various attention mechanisms and ideas from positional encodings as well as deeply-supervised networks \cite{LeeXGZT2015}, in combination with a performant encoder which has proven its applicability for HTR tasks. The proposed model sets a solid foundation for Seq2Seq models in HTR and allows for future research on the direct integration of external memory modules as well as language models in the decoder, which has not been done for purely CTC-based architectures. 

The paper is structured as follows: Sec.~\ref{sec:rel_work} reviews relevant works for HTR and related topics. Sec.~\ref{sec:method} describes the proposed methodology and Sec.~\ref{sec:implementation} explains the implementation details. Afterwards, Sec.~\ref{sec:exp} presents different experimental setups, provides results and compares against the state-of-the-art. Finally, Sec.~\ref{sec:conclusion} concludes and contemplates future work.

\section{Related Work}\label{sec:rel_work}
The main challenge in sequence learning tasks is to find an appropriate alignment between input and output sequences of variable length. For unconstrained HTR, one needs to identify the correct characters at each time step without any prior knowledge about the alignment between the image pixels and the target characters. The two major methods in deep learning that overcome this problem are CTC- and Seq2Seq-based approaches, the latter usually with attention.

CTC-based models compute a probability distribution over all possible output sequences, given an input sequence. They do so by dividing the input sequence into frames and emitting, for each frame, the likelihood of each character of the target alphabet.\comment{expanded by an artificial blank character.} The probability distribution can be used to infer the actual output greedily, either by taking the most likely character at each time step or using beam search.

In the last years, the main CTC-based architectures for HTR were Long Short-Term Memory (LSTM) \cite{HochreiterS1997} networks, in many cases using the bidirectional variant (BLSTM) \cite{GravesSchmidhuber2005}. 
Others, like \comment{\cite{GravesSchmidhuber2008, BlucheLKMBK2014}} \cite{PhamKL2013, VoigtlaenderDN2016}, proposed deep networks based on Multidimensional LSTMs (MDLSTMs) \cite{GravesFS2007}, that exploit the two-dimensional nature of handwritten images. The current state-of-the-art in many text recognition tasks additionally integrate CNNs for an improved low-level feature extraction prior to the recurrent layers. This approach is applied to offline HTR by \cite{Puigcerver2017}.
Recently there have been developments towards fully convolutional architectures, i.e. recurrence-free, as in \cite{YousefHS2018}, that reach competitive HTR performance.


Alternatively, the idea of Seq2Seq architectures that follow the encoder-decoder framework, is to decouple the decoding from the feature extraction. The models consist of two main parts. First, an encoder reads and builds a feature representation of the input sequence, then a decoder emits the output sequence one token at a time. Usually an attention mechanism is employed by the decoder to gather context information and search for relevant parts of the encoded features.

Without any line segmentation, \cite{Bluche2016} uses MDLSTMs for full paragraph recognition. Many authors combine a CNN as a feature extractor with an RNN encoder and unidirectional decoder for use in HTR. In particular, \cite{PoulosValle2017} and \cite{ChowdhuryVig2018} use BLSTMs for the recurrent encoder. Some works, like \cite{SueirasRSV2018, KangTRVFR2018}, use similar architectures, but limit their work on recognizing isolated handwritten words. A bidirectional decoder is incorporated in \cite{DoetschZN2016}, by integrating a length estimation procedure.

\section{Proposed Methodology}\label{sec:method}
\begin{figure}[tb]
	\centering
	\includegraphics[page=4,width=0.48\textwidth]{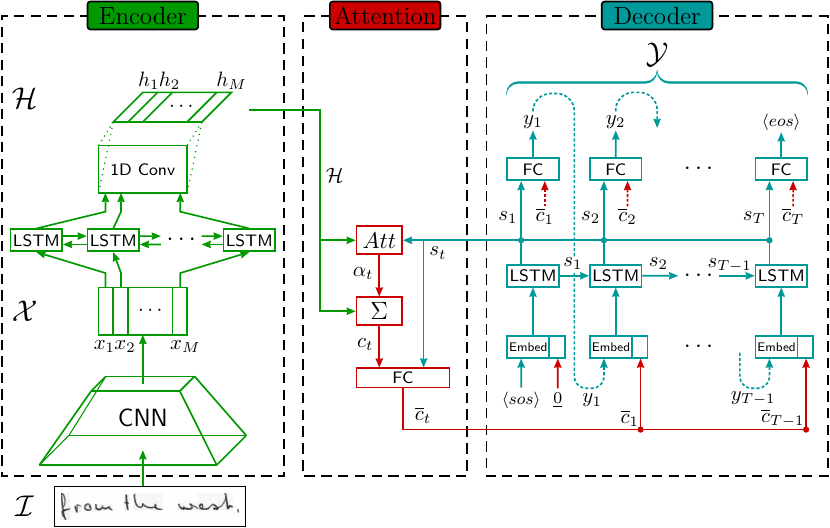}
	\caption{\textbf{General architecture of the attention-based Seq2Seq model.} The encoder converts an input image $\mathcal{I}$ into a sequence of constant feature vectors $\mathcal{H}$. The decoder emits the output sequence $\mathcal{Y}$ one character at a time. At each time step $t$, it employs an attention mechanism $Att$ to generate a context vector $\overline{c}_t$ based on the encoded feature vectors and the time-dependant decoder hidden state $s_t$. This context vector is used to produce the decoders output $y_t$ for the current time step. A concatenation of the context vector and the embedded output serves as the decoders next input.}
	\label{fig:seq2seq}
\vspace{-4mm}
\end{figure}

Our \comment{attention-based} Seq2Seq model follows the standard encoder-decoder framework with attention\comment{, which enables us to map variable length input sequences into variable length output sequences}. 
The model consists of three main parts: an encoder that combines a CNN as a generic feature extractor with recurrent layers to introduce temporal context in the feature representation, a decoder that utilizes a recurrent layer to interpret those features and an attention mechanism that enables the decoder to focus on the most relevant encoded features at each decoding time step. In the following, these three components are described in more detail. A general overview of the architecture is shown in Fig.~\ref{fig:seq2seq}.

\subsection{Encoder}
The encoder starts with a deep CNN, as those have proven to be effective in extracting visual features from images. The CNN converts segmented text line images of variable length and fixed height into a sequence of visual feature vectors. It comprises three successive convolutional layers with interleaved pooling layers and reduces the spatial dimensions of the image while simultaneously increasing the representative depth of the feature vectors.

Next, we introduce a deep RNN that reads the sequence of convolutional features and tries to encode temporal context between them. 
We use BLSTM layers, that combine two LSTM layers which process the sequence in opposite directions to encode both forward and backward dependencies, in order to capture the natural relationship of written text. \comment{, where neighborhoods of words are correlated.} The encoder-RNN comprises three such BLSTM layers.

Formally, the encoder-CNN processes an input image $\mathcal{I}$ and transfers it into an intermediate-level feature map $\mathcal{X}$, which can be thought of as a sequence of column vectors $\mathcal{X}=(x_1,\dots,x_M)$, where $M$ is the subsampled input sequence length. 
This sequence is then processed by the BLSTM layers and we get the combined state sequence \comment{$\mathcal{H'} = (h'_1,\dots,h'_M)$} as the sum of the forward and backward hidden states. A 1D convolutional layer over the state sequence of the third BLSTM layer produces the final encoded feature map $\mathcal{H} = (h_1,\dots,h_M)$, that models the visual information of the input image on the one hand and the temporal context of the sequence on the other hand. Note that a suitable choice for the convolutional output layer parameters makes this encoder fit the framework of a purely CTC-based architecture. This can be beneficial, as it preserves an interpretable and usable encoder representation. Furthermore, it opens up possibilities to integrate CTC-pretrained model weights into the system to potentially increase the models performance.

\subsection{Decoder}
We implement the decoder-RNN as a unidirectional LSTM layer, as its purpose is to generate the target character sequence present in the image. At each time step $t$, the decoder computes a probability distribution over the alphabet of possible characters and predicts the most probable character $y_t$, conditioned on its own previous predictions $(y_1,\dots,y_{t-1})$ and some context vector $c_t$ which contains information from the encoded features $\mathcal{H}$. Basically, it defines a probability over the output sequence $\mathcal{Y}=(y_1,\dots,y_T)$, by decomposing the joint probability into the ordered conditionals $p(\mathcal{Y}) = \prod_{t=1}^T p(y_t\mid y_1, \dots, y_{t-1}, c_t)$, where $T$ is the output sequence length. 
Though, the conditions on $(y_1,\dots,y_{t-2})$ are modeled implicitly, that is, each conditional is modeled as $p(y_t\mid y_1, \dots, y_{t-1}, c_t) = \text{softmax}(f(y_{t-1}, s_{t-1}, c_t))$, where $f$ represents the LSTM and $s_{t-1}$ its previous hidden state. Particularly, the decoder has to gather information about the history of the output sequence and save it in its internal memory state. Contrary to popular standards, we initialize the decoder LSTM with a zero state instead of the final state of the encoder BLSTM. Although the basic setup, which uses a fixed context vector $c_t = c = h_M$, is able to handle a sequence-to-sequence mapping, it is apparent that the context vector $c$ forms a bottleneck as it is the only link between the encoder and the decoder. Instead, the decoder employs an attention mechanism in order to focus on the most relevant part of the encoded feature representation at each decoding time step.

\subsection{Attention mechanism}
Attention mechanisms extend the standard encoder-decoder framework, by dynamically modifying the context vector at each time step based on some similarity of the decoder hidden state $s_t$ with the encoded features $\mathcal{H}$ for a particular input sequence. In the general case, the context vector $c_t$ at decoding time step $t$ is given by a weighted sum over the sequence of encoded feature vectors
\begin{align}
c_t = \sum_{j=1}^M \alpha_{t,j}h_j. \label{eq:context}
\end{align}
The attention weights $\alpha_{t,j}$ are formulated as normalized attention scores and are subject to the particular attention function used
\begin{align}
\alpha_{t,j}=Att(s_t,h_j, \alpha_{t-1}).\label{eq:att}
\end{align}
This way, the decoder can learn local correspondence between the input and output sequences in conjunction with a global context. We distinguish six kinds of attention mechanisms.

\subsubsection{Content-based attention}
Dropping the attention vector $\alpha_{t-1}$ from \eqref{eq:att} results in purely content-based attention, in which the decoder is only concerned about what encoded features fit its hidden state the best. Examples include Bahdanau-style attention \cite{BahdanauCB2014} or Luong-style attention \cite{LuongPM2015}, where the attention weights are usually computed as the normalized similarity scores
\begin{align}
e_{t,j} &= \begin{cases}
			s_t^\mathsf{T}W_h h_j & \text{Luong} \\
			v^\mathsf{T}\tanh(W_s s_t + W_h h_j + b) & \text{Bahdanau}
		 \label{eq:score}
		 \end{cases} \\
\alpha_{t,j} &= \text{softmax}(e_{t,j}),
\end{align}
with trainable parameters $W_h, W_s, v, b$ and an elementwise tanh function. The drawback of this approach is that similar elements of $\mathcal{H}$ are scored equally regardless of their position in the sequence, so that the decoder will never be able to detect the difference between multiple feature representations of the same character in different positions. This issue is partially alleviated by the encoder-BLSTM, which is able to encode temporal context into the feature vectors $h_j$, however, this is only possible to a limited extent.

\subsubsection{Penalized attention}
In order to not attend to the same encoded features over and over again, one can try to penalize feature vectors that have obtained high attention scores in past decoding steps. This kind of penalized attention \cite{SankaranMAI2016} defines temporal scores $e'_{t,j}$ that are then used to compute the attention weights.
\vspace{-4mm}
\begin{align}
e'_{t,j} = \begin{cases}
			\exp(e_{t,j}) & \text{if } t = 1 \\
			\frac{\exp(e_{t,j})}{\sum_{i=1}^{t-1}\exp(e_{i,j})} & \text{otherwise}
		   \end{cases} 
\end{align}

\subsubsection{Location-based attention}
Alternatively, dropping the encoded feature vectors $h_j$ from \eqref{eq:att} yields purely location-based attention like in \cite{Graves2013}. In the HTR scenario, these types of attention mechanisms would have to predict the distance between characters, as well as the relevant context size, using $s_t$ only, which we think is not quite feasible.

\subsubsection{Monotonic attention}
A more natural candidate for HTR seems to be attention that incorporates both content and location information. The latter can be done implicitly, by forcing an alignment that fits the underlying task. In the case of HTR, one can argue that a monotonic alignment between the input image and the output character sequence seems natural. Such an alignment can be forced by using a Monotonic Attention mechanism \cite{RaffelLLWE2017}. The idea is to process the memory in a left-to-right manner, computing "choosing probabilites" for each vector of $\mathcal{H}$ based on its similarity score and to stop as soon as a certain entry $h_j$ is sampled. This entry would serve as the context vector for that particular time step $c_t = h_j$. In practice, to enable the use of backpropagation, training is done with respect to the expected value of $c_t$ and we keep this procedure even for inference. 

\subsubsection{Chunkwise attention}
However, the hard monotonicity constraint also limits the expressivity of the model compared to purely content-based attention, since the context for a particular timestep is narrow and the input-output alignment must be strictly monotonic. A compromise between soft attention and hard monotonicity is Monotonic Chunkwise Attention (MoChA) \cite{ChiuR2018}, which adaptively splits the input sequence into small chunks over which soft attention is performed. The attention mechanism follows the hard monotonic attention process in order to determine a particular memory entry $h_j$. However, instead of setting $c_t=h_j$, the model is allowed to perform soft attention over a length-$w$ window of memory entries preceding and including $h_j$. This allows for reordering of the windows' memory entries \comment{$h_{j-w+1},\dots,h_j$}, i.e. non-monotonic alignments. Analogous to monotonic attention, training is done using the expected value of $c_t$ and we keep this procedure even for inference.

\subsubsection{Hybrid attention}
A different approach explicitly models location awareness, by incorporating the attention weights of the previous time step into the attention mechanism, as proposed in \cite{ChorowskiBSCB2015}. We use a 1D convolutional layer to extract feature vectors $f_{t,j}$ for every position $j$ of the previous alignment $\alpha_{t-1}$ 
and use these vectors to modify the scoring function, e.g. for Bahdanau score \eqref{eq:score}:
\begin{align}
e_{t,j} &= v^\mathsf{T}\tanh(W_s s_t + W_h h_j + W_f f_{t,j} + b).
\end{align}
The computation of the attention weights $\alpha_{t,j}$ and the context vector $c_t$ remains as usual. This way, the location-aware approach can be combined with either of the previously presented attention mechanisms, which we call Hybrid Attention.

\subsection{Positional encoding}
Positional encodings inject some information about the relative or absolute position of the tokens in the sequence. They are added to the final feature representation of the encoder. In this work, we evaluate two different approaches.

Firstly, we use fixed sine and cosine functions of different frequencies \cite{VaswaniSPUJGKP2017}.
Secondly, we use an embedding matrix to retrieve learned positional encodings \cite{GehringAGYD2017} for each of the encoded feature vectors.

\subsection{Training}
As a combination of differentiable neural modules, our model can be trained end-to-end in a supervised manner via the backpropagation algorithm. The common approach is to train the encoder-decoder jointly by minimizing the cross-entropy loss $\mathcal{L}_{ce}=\sum_{t=1}^T-\log p(y_t=g_t\mid y_1, \dots, y_{t-1},  c_t)$, where $g_t$ is the target token of time step $t$.

We also want to investigate the idea of introducing local objectives for intermediate layers, one of the general ideas of deeply-supervised networks. Therefore, we integrate the standard CTC loss $\mathcal{L}_{ctc}$, if applicable, to further tune the encoder. This kind of multi-task learning approach also showed performance improvements in end-to-end speech recognition \cite{KimHW2016}. The resulting hybrid loss is formulated as a convex combination of both training losses
\begin{align}
\mathcal{L}=\lambda\mathcal{L}_{ctc}+(1-\lambda)\mathcal{L}_{ce},\,\lambda\in[0,1].
\end{align}
\section{Implementation details}\label{sec:implementation}
We performed grid and random search to select model hyperparameters. \comment{, like the dimensions of intermediate layers and embedding matrices, the intial learning rate, the optimization strategies or the form and value of gradient clipping.} Our final base model was evaluated in different setups and has the following properties.

\subsection{Encoder}
The encoder is a three-layer CNN with interleaved max-pooling layers, followed by three BLSTM layers: \comment{A general overview is depicted in the following.}
\begin{align*}
\mathcal{I}&\rightarrow C_{6\times4}^{4\times2}[8]\rightarrow C_{6\times4}^{1\times1}[32]\rightarrow P_{4\times2}^{4\times2}\rightarrow C_{3\times3}^{1\times1}[64]\rightarrow P_{1\times2}^{1\times2} \\
&\rightarrow R[256]\rightarrow R[256]\rightarrow R[256]\rightarrow C_{256\times1}^{1\times1}[o]\rightarrow\mathcal{H}
\end{align*}
where $C$, $P$ and $R$ represent a convolutional, a max-pooling and a BLSTM layer respectively. The sub- and superscript in $C/P_{k_y\times k_x}^{s_y\times s_x}$ desribe the size of the kernel $k$ and strides $s$ along both dimensions (width $x$, height $y$). The depth of the feature map is shown in square brackets, i.e. the number of filters for a convolutional layer or the number of hidden units in each LSTM. Choosing the number of output channels $o$ to match the number of character channels (extended by a blank channel), ensures the applicability of the encoder output and allows for an easy integration of the hybrid loss. Each convolutional layer uses leaky rectified linear units \cite{MaasHN2013} as activation functions, with weights initialized using Xavier initialization \cite{GlorotB2010}. Dropout was applied to each BLSTM output at train time with a probability of \percent{50}.

\subsection{Decoder}
The decoder is a single unidirectional LSTM with 256 hidden units and a dropout probability of \percent{50} at train time. Increasing the depth of the decoder did not improve its predictive power in our experiments. We use 64-dimensional learned character embeddings, to feed in the output of the decoder at the next time step. The decoder employs an attention mechanism to compute an attention vector and  uses a fully-connected layer to compute the logits. Generating the final prediction in inference is done using beam search with a beam width of 16 for all our experiments.

\subsection{Attention mechanism}\label{subsec:att}
We evaluate six different attention mechanisms: content-based, penalized and location-based attention, monotonic and chunkwise attention, as well as hybrid attention. We use Bahdanau-style scoring functions, since Luong-style scoring functions performed worse across some prelimenary experiments. We further differentiate between the standard $v^\mathsf{T}\tanh(W_s s_t + W_h h_j + b)$ and normalized $g\frac{v^\mathsf{T}}{\lVert v\rVert}\tanh(W_s s_t + W_h h_j + b) + r$ forms of the scoring function. Chunkwise attention uses a window size of 3, since increasing the windows produced noisier results. Hybrid attention uses 
a 1D convolutional layer of the form $C_{7\times1}^{1\times1}[20]$. The size of the attention layers that transform the decoder hidden state $s_t$, the encoded feature vector $h_j$ and the location vectors $f_{t,j}$ into matching dimensions is fixed to 128. We additionally feed the context vector $c_t$ and decoder hidden state $s_t$ into a fully-connected layer with 128 units to generate attention $\overline{c}_t$ at each time step (see Fig.~\ref{fig:seq2seq}).

\subsection{Training}
Training is done using the ADAM optimizer \cite{KingmaB2014} with a mini-batch size of 16 for 200 runs (epochs) with 8192 training samples each. We use an initial learning rate of 0.001 and apply a cosine-like decay in the final 50 epochs to finetune the parameters of the network. Hybrid loss training uses an equal weighting with $\lambda=0.5$. Experiments with adaptive weightings, e.g. starting with a big $\lambda$ to quickly adapt the encoder and letting it decay linearly to increase the training signal for the decoder at later stages of the training, did not provide any noticeable benefits. The target sequences are expanded by artifical $\langle sos\rangle$ and $\langle eos\rangle$ tokens, representing the start and end of the sequence respectively. \comment{The $\langle sos\rangle$ token serves as the initial input to the decoder and once the decoder outputs an $\langle eos\rangle$ token, the decoding process is finished.} At train time, teacher forcing ensures that the decoder sees the correct character of the previous timestep, i.e. we feed in the embedded target token instead of the decoders last output. Nevertheless, we integrate some teacher noise to improve generalization, where this mechanism is skipped with a probability of \percent{10}. In this case a character from the decoders output probability distribution is sampled and its embedding is fed in as the next input.

\subsection{Preprocessing and data augmentation}
Preprocessing of the text line images includes a contrast normalization without any binarization of the image, a skew correction and a slant normalizer. \comment{using the centre-of-gravity-line of a text line.} Additionally, all images are scaled to a fixed height of 64 pixel while maintaining their aspect ratio. This ensures that all vectors in the feature sequence conform to the same dimensionality without putting any restrictions on the sequence length. \comment{Neural networks typically require large amounts of data to learn and generalize.} To artificially increase the amount of training data, we augment the existing preprocessed images by applying minor alterations to them. To simulate naturally occuring variations in handwritten text line images, we combine dilation, erosion and grid-like distortions \cite{WigingtonSDBPC2017}. 
These methods are applied to the original line image randomly with an independent probability of \percent{50}.

%
\section{Experimental Evaluation}\label{sec:exp}
This section is divided into four subsections. First, we present the data sets on which we evaluated our models. In Sec.~\ref{subsec:fixed} we investigate the performance of the Seq2Seq architecture for a pretrained encoder with fixed parameters, i.e. the encoder parameters are not adjusted during training of the decoder. Such a setup ensures an interpretable and usable encoder representation. Since the results in this setup are not satisfying at all, Sec.~\ref{subsec:hybrid} provides results of a hybrid loss approach, inspired by the recent success of deeply-supervised networks. The first two experimental setups help us to determine a general setup concerning the attention mechanism and positional encoding. Finally, in Sec.~\ref{subsec:final} the final results achieved by the investigated architecture are presented and compared to the current state-of-the-art. For the first two experimental subsections, we limit ourselves to the IAM data set and rank our systems based on the validation error. Of note, since we train our systems for a fixed number of epochs and do not perform any early stopping based on the validation error, this is a valid procedure. Evaluation is made by comparing the estimated transcription of the model with the target character sequence. It should be noted, that we do not use a language model to improve the final prediction. Since we focus on the raw performance of the optical models, we measure the Character Error Rate (CER), i.e. the edit distance (Levenshtein distance) normalized by the number of characters in the target, instead of the Word Error Rate.

\subsection{Data sets}

The \textbf{IAM Handwriting Database} \cite{MartiB2002} contains forms of unconstrained handwritten text.
We use the so called Aachen's partition\footnote{taken from \url{https://github.com/jpuigcerver/Laia/tree/master/egs/iam}} (IAM-A) of the data set. 
This is not the official partition, but it is widely used for HTR experiments. Other partitions, which we call IAM-B and IAM-C\footnote{\url{http://www.fki.inf.unibe.ch/databases/iam-handwriting-database}}, differ slightly in the number of included text lines.
%


The \textbf{ICFHR2016 READ data set (Bozen)} \cite{Bozen2016} is based on the collection of the state archive of Bozen. It arised from the European Union’s Horizon 2020 READ project and is openly available. The data set consists of a subset of documents from the Ratsprotokolle collection composed of minutes of the council meetings held from 1470 to 1805. This data set was the basis for a HTR competition at the ICFHR2016 \cite{SanchezRTV2016}.  
%

The \textbf{StAZH data set} is based on the collection of the state archive of the canton of Zurich. This is an internal data set of the European Union’s Horizon 2020 READ project and thus not openly available yet, but we included it for additonal results on historical handwritings.
The documents contain resolutions and enactments of the cabinet as well as the parliament of the canton of Zurich from 1803 to 1882.

The data sets are divided into training, validation and test sets, with the number of segmented lines depicted in Tab.~\ref{tab:datasets}.

\begin{table}[tb]
	\renewcommand{\arraystretch}{1.2}
	\centering
	\caption{\textbf{Data sets used for evaluation --} The number of segmented lines for each subset is shown.}
	\label{tab:datasets}
	\centering
	\begin{tabular}{|c||c|c|c|c|c|}
		\hhline{-||-----}
		\multirow{2}{*}{Set}& \multicolumn{5}{c|}{Number of segmented lines}\\
		& IAM-A & IAM-B & IAM-C & Bozen & StAZH \\
		\hhline{=::=:=:=:=:=}
		Training & $6161$ & $6482$ & $6161$ & $8367$ & $12628$ \\
		Validation & $966$ & $976$ & $940$ & $1043$ & $1624$ \\
		Test & $2915$ & $2915$ & $1861$ & $1140$ & $1650$ \\
		\hhline{-||-----}
	\end{tabular}
\end{table}


\subsection{Fixed Encoder}\label{subsec:fixed}
In a first experimental setup we compare different attention mechanisms and possibilities to directly encode the positional information in the encoder output. This is done for a fixed encoder, i.e. we feed the weights of a CTC-pretrained model into the encoder part of the Seq2Seq model and fixate them during training. Consequently, the encoder CER remains constant in those experiments. Tab.~\ref{tab:enc_pre} lists the CER performance for our encoders trained as a stand-alone system, using solely the CTC loss and a greedy output. We investigate the different attention mechanisms and positional encoding methods (sinusoid PE$_s$, learned PE$_l$) for our Seq2Seq model, which were presented in Sec.~\ref{sec:method}. For these experiments, we trained each setup just once, since the results can not compete with the stand-alone encoder, as shown by Tab.~\ref{tab:s2s_pre}.

\begin{table}[tb]
	\renewcommand{\arraystretch}{1.2}
	\centering
	\caption{\textbf{Results for a stand-alone encoder --} The validation CER on the respective data set is shown.}
	\label{tab:enc_pre}
	\centering
	\begin{tabular}{c||c|c|c|}
		\hhline{~|---}
		\multicolumn{1}{c|}{} & \multicolumn{3}{c|}{Validation CER in \%}\\
		\multicolumn{1}{c|}{} & IAM & Bozen & StAZH \\
		\hhline{~|-|-|-}\noalign{\vspace*{\doublerulesep}}
		\hhline{-||-|-|-}		
		\multicolumn{1}{|c||}{stand-alone encoder}		
		 & $3.53$ & $6.29$ & $3.10$ \\
		\hhline{-||---}
	\end{tabular}
\end{table}


\begin{table}[tb]
	\renewcommand{\arraystretch}{1.2}
	\centering
	\caption{\textbf{Results for a fixed encoder --} The validation CER for the encoder and decoder are shown for different attention mechanisms and positional encodings on the IAM data set.}
	\label{tab:s2s_pre}
	\centering
	\begin{tabular}{|l|c||c|c|}
		\hhline{--||--}
		\multirow{2}{*}{Attention}&\multirow{2}{*}{Encoding}& \multicolumn{2}{c|}{Validation CER in \%}\\
		&& Encoder & Decoder  \\
		\hhline{=:=::=:=}
		content & -- & $3.53$ & $64.92$ \\
		content & PE$_s$ & $3.53$ & $20.43$ \\
		content & PE$_l$ & $3.53$ & $11.51$ \\
		\hhline{--||--}
		location & -- & $3.53$ & $100.00$ \\
		\hhline{--||--}
		penalized & -- & $3.53$ & $80.83$ \\
		\hhline{--||--}
		monotonic & -- & $3.53$ & $5.61$ \\
		\hhline{--||--}
		chunkwise & -- & $3.53$ & $6.48$ \\
		chunkwise & PE$_l$ & $3.53$ & $4.24$ \\
		\hhline{--||--}
	\end{tabular}
\vspace{-3mm}
\end{table}

Because of the purely content-aware attention mechanism in the first setup, the CER of \percent{64.92} is in the range of expectation. The decoder is not aware of where the content it attends to is positioned in the sequence and because the encoder is fixed, the recurrent weights can not adapt accordingly. Hence, the decoder tends to arrange correct or similar text snippets in a quite arbitrary order and repeats certain parts multiple times.


Purely location-based attention is not capable to yield the necessary information for the decoder as expected. Similarly, penalizing features that the decoder attended to in previous time steps, in order to reduce repetitions, does not improve overall performance. One can alleviate these problems to a certain degree by encoding the positional information or by enforcing a monotonic alignment in the attention mechanism. However, the obtained results are still worse than those achieved by the stand-alone encoder model. Systems using positional encodings in combination with content-based attention have problems \comment{emitting the $\langle eos\rangle$ token, i.e.} predicting the end of the sequence.


\subsection{Hybrid Loss}\label{subsec:hybrid}
Because the results obtained with a fixed encoder are not satisfying at all, we allow the adaptation of the encoder parameters in this setup and use a hybrid loss with $\lambda=0.5$ as training signal. This is inspired by deeply-supervised networks and should retain the interpretable encoder output. The intuition is, that the encoder output maintains its text-predictive power and in addition some meaningful positional information is overlayed and utilized by the decoder. The equal weighting of the cross-entropy and CTC loss is a reasonable choice, because both losses are of same magnitude. All experiments were performed three to five times to reduce the influence of training noise and to average out any statistical outliers. The results in Tab.~\ref{tab:s2s_hybrid} show that this setup enables the decoder to produce good results and that the end-to-end training even improves the encoder performance in comparison to the stand-alone model - if only marginally. We observe that positional encodings do not add any benefit in this setup and that strictly monotonic attention is on par with monotonic chunkwise attention. Penalized attention is still not competitive and although location-based attention performed not as consistent as others, \comment{attention mechanisms} the system could benefit from additional location awareness. 

\begin{table}[tb]
	\renewcommand{\arraystretch}{1.2}
	\centering
	\caption{\textbf{Results for hybrid loss --} The validation CER for the encoder and decoder are shown for different attention mechanisms and positional encodings on the IAM data set.}
	\label{tab:s2s_hybrid}
	\centering
	\begin{tabular}{|l|c||c|c|}
		\hhline{--||--}
		\multirow{2}{*}{Attention}&\multirow{2}{*}{Encoding}& \multicolumn{2}{c|}{Validation CER: mean [min, max] in \%}\\
		&& Encoder & Decoder  \\
		\hhline{==::==}
		content & -- & $3.55\,[3.49, 3.62]$ & $3.58\,[3.52, 3.61]$ \\
		content & PE$_s$ & $3.54\,[3.44, 3.61]$ & $3.65\,[3.54, 3.75]$ \\
		content & PE$_l$ & $3.57\,[3.54, 3.61]$ & $3.62\,[3.59, 3.65]$ \\
		\hhline{--||--}
		location & -- & $3.49\,[3.44, 3.56]$ & $4.18\,[3.62, 4.81]$ \\
		\hhline{--||--}
		penalized & -- & $36.3\,[11.6, 69.8]$ & $58.6\,[25.8, 83.4]$ \\
		\hhline{--||--}
		monotonic & -- & $3.53\,[3.51, 3.57]$ & $3.54\,[3.50, 3.56]$ \\
		monotonic & PE$_s$ & $3.55\,[3.54, 3.56]$ & $3.52\,[3.49, 3.55]$ \\
		monotonic & PE$_l$ & $3.52\,[3.49, 3.61]$ & $3.55\,[3.52, 3.57]$ \\
		\hhline{--||--}
		chunkwise & -- & $3.44\,[3.38, 3.48]$ & $3.46\,[3.38, 3.57]$ \\
		chunkwise & PE$_s$ & $3.50\,[3.47, 3.55]$ & $3.55\,[3.50, 3.62]$ \\
		chunkwise & PE$_l$ & $3.52\,[3.49, 3.57]$ & $3.54\,[3.48, 3.60]$ \\		
		\hhline{--||--}
	\end{tabular}
\vspace{-1mm}
\end{table}

\begin{table}[tb]
\vspace{-1mm}
	\renewcommand{\arraystretch}{1.2}
	\centering
	\caption{\textbf{Results for hybrid attention --} The validation CER for the encoder and decoder are shown for miscellaneous settings on the IAM data set.}
	\label{tab:s2s_misc}
	\centering
	\begin{tabular}{|l|l||c|c|}
		\hhline{--||--}
		\multirow{2}{*}{Attention}&\multirow{2}{*}{Misc}& \multicolumn{2}{c|}{Validation CER: mean [min, max] in \%}\\
		&& Encoder & Decoder  \\
		\hhline{==::==}
		hy-mono & -- & $3.53\,[3.50, 3.58]$ & $3.50\,[3.50, 3.50]$ \\
		hy-mono & N & $3.50\,[3.46, 3.55]$ & $3.46\,[3.45, 3.47]$ \\
		hy-mono & N + C & $3.50\,[3.45, 3.53]$ & $\mathbf{3.41\,[3.40, 3.42]}$ \\
		hy-mono & N + C + PE$_s$ & $3.55\,[3.52, 3.56]$ & $3.55\,[3.50, 3.62]$ \\
		hy-mono & N + C + PE$_l$ & $3.46\,[3.43, 3.50]$ & $3.45\,[3.44, 3.46]$ \\
		hy-mono & N + C + $\mathcal{L}^{s}_{ce}$ & $100\,[100, 100]$ & $3.50\,[3.45, 3.54]$ \\
		hy-mono & N + C + $\mathcal{L}^{p}_{ce}$ & $52.9\,[45.0, 67.4]$ & $3.47\,[3.40, 3.51]$ \\
		\hhline{--||--}
		hy-chunk & N & $3.47\,[3.46, 3.48]$ & $3.45\,[3.43, 3.48]$ \\
		hy-chunk & N + C & $3.46\,[3.44, 3.47]$ & $3.48\,[3.46, 3.51]$ \\
		\hhline{--||--}
	\end{tabular}
\vspace{-3mm}
\end{table}

\begin{table*}[tb]
	\renewcommand{\arraystretch}{1.2}
	\centering
	\caption{\textbf{Results for final model --} The validation and test CER for the decoder as well as the encoder are shown on all data sets. In addition to the standard training set, a second track includes the validation data during training.}
	\label{tab:s2s_final}
	\centering
	\begin{tabular}{|l|l||c|c|c|c|}
		\hhline{--||----}
		\multirow{2}{*}{Data set} & \multirow{2}{*}{Training set} & \multicolumn{2}{c|}{Encoder CER: mean [min, max] in \%} & \multicolumn{2}{c|}{Decoder CER: mean [min, max] in \%} \\
		&& validation set & test set & validation set & test set \\
		\hhline{==::====}
		\multirow{2}{*}{IAM} & training & $3.50\,[3.45, 3.53]$ & $5.31\,[5.30, 5.32]$ & $3.41\,[3.40, 3.42]$ & $5.24\,[5.21, 5.26]$ \\
		& training + validation & -- & $4.96\,[4.93, 4.99]$ & -- & $4.87\,[4.84, 4.89]$ \\
		\hhline{--||----}
		\multirow{2}{*}{Bozen} & training & $6.16\,[6.03, 6.28]$ & $5.04\,[5.02, 5.08]$ & $6.00\,[5.87, 6.10]$ & $4.99\,[4.94, 5.07]$ \\
		& training + validation & -- & $4.76\,[4.68, 4.88]$ & -- & $4.66\,[4.60, 4.71]$ \\
		\hhline{--||----}
		\multirow{2}{*}{StAZH} & training & $3.00\,[2.98, 3.05]$ & $2.87\,[2.84, 2.92]$ & $2.87\,[2.83, 2.94]$ & $2.78\,[2.74, 2.83]$ \\
		& training + validation & -- & $2.58\,[2.55, 2.59]$ & -- & $2.48\,[2.45, 2.49]$ \\
		\hhline{--||----}
	\end{tabular}
\end{table*}

Going forward, we therefore focused on hybrid monotonic (hy-mono) and hybrid monotonic chunkwise (hy-chunk) attention. We also experimented with some miscellaneous settings like the normalized form of the attention scoring function (N) (see Sec.~\ref{subsec:att}) and gradient clipping (C), \comment{differing window sizes for the chunkwise attention part (W$_{size}$), adding pre-sigmoid noise to the monotonic attention part} where we normalize the gradient tensors by their $L_2$-norm if it exceeds $4.0$ in value. 
For the most promising architecture we reintroduced the two variants of positional encodings and, for reference, trained seperate models using solely the cross-entropy loss: one from scratch ($\mathcal{L}^{s}_{ce}$) and one with a pretrained encoder ($\mathcal{L}^{p}_{ce}$). The results for these miscellaneous experiments are shown in Tab.~\ref{tab:s2s_misc}. Hybrid monotonic attention marginally outperforms hybrid chunkwise attention when using the normalized scoring function and clipping the gradients. Learned positional encodings work better than fixed ones, but still do not improve the overall performance. Dropping the hybrid loss makes the encoder lose its text-predictive power as expected. Nevertheless, both the model trained from scratch aswell as the model with a pretrained encoder show, that the hybrid loss approach is not necessary for HTR purposes, if one is not interested in preserving an interpretable encoder output.

\subsection{Final evaluation}\label{subsec:final}
Based on the previous experimental results, we chose hybrid monotonic attention with the normalized Bahdanau scoring function in combination with gradient clipping and the implementation details of Sec.~\ref{sec:implementation} as our final architecture. We evaluate the model on both the validation and the test set. Since more training data is beneficial, we additionally provide results for models which are trained on the training and validation set, which is a common approach in real-world scenarios, where training data is scarce. We show our final evaluation in Tab.~\ref{tab:s2s_final}.

Although the performance difference between the encoder (greedy CTC output) and the decoder is only marginal, our Seq2Seq model is generally not bound to an encoder that conforms to the CTC setup (e.g. the output dimension and limitations to the sequence length). This way we remain flexible in relation to subsampling the input image and the alignment between the input and output sequence.

Comparative results on the Bozen and IAM data sets are provided in Tab.~\ref{tab:s2s_bozen} and Tab.~\ref{tab:s2s_iam} respectively. In both cases we indicate what kind of post-processing was applied, like the use of a language model (LM) or a lexicon (LX).

On the Bozen test set we achieve an average CER of \percent{4.66}, which beats the top participant of the corresponding competition, although they used a 10-gram character-based LM. Furthermore, we achieve an average CER of \percent{4.87} on the IAM test set. A direct comparison to the state-of-the-art is not possible, because most rely on the use of additional language resources for decoding. Instead, comparing our results to the direct errors of the visual models shows, that we reach very good results on the character level. In fact, we are on par with the state-of-the-art results without post-processing and make significant improvements over the latest works that use a similar Seq2Seq approach. Particularly, we make a relative improvement of about \percent{29} over \cite{KangTRVFR2018}, although textline CERs are typically higher than word-level ones. Integrating suitable LMs will be part of future work. For instance, the results from \cite{VoigtlaenderDN2016} were achieved by combining a smoothed word-based trigram with a 10-gram character-based LM.

\begin{table}[tb]
	\renewcommand{\arraystretch}{1.2}
	\centering
	\caption{\textbf{Comparative results Bozen --} The test CER on the Bozen data set is shown, as well as the kind of post-processing.}
	\label{tab:s2s_bozen}
	\centering
	\begin{tabular}{|ll||c|}
		\hhline{--||-}
		Architecture & Method & Test CER in \% \\
		\hhline{==::=}
		CNN-MDLSTM [RWTH] & CTC + LM & $4.8$ \\
		CNN-RNN [BYU] & CTC & $5.1$ \\
		CNN-MDLSTM [A2IA] & CTC + LM & $5.4$ \\
		BLSTM [LITIS] & CTC + LX & $7.3$ \\
		BLSTM [ParisTech] & CTC & $18.5$ \\
		\hhline{--||-}
		Ours & Seq2Seq & $\mathbf{4.66}$ \\
		\hhline{--||-}
	\end{tabular}
\end{table}

\begin{table}[tb]
\centering
\begin{threeparttable}[tb]
	\renewcommand{\arraystretch}{1.2}
	\caption{\textbf{Comparative results IAM --} The test CER on the IAM data set is shown, as well as the kind of post-processing and underlying partition.}
	\label{tab:s2s_iam}
	\begin{tabular}{|ll||c|}
		\hhline{--||-}
		Architecture & Method & Test CER in \% \\
		\hhline{==::=}
		\multirow{2}{*}{CNN-MDLSTM \cite{PhamKL2013}} & CTC\tnote{1} & $10.8$ \\
		& CTC\tnote{1} $\,$+ LM + LX & $5.1$ \\[1mm]
		CNN-MDLSTM \cite{VoigtlaenderDN2016} & CTC\tnote{1} $\,$+ LM & $\mathbf{3.5}$ \\[1mm]
		\multirow{2}{*}{CNN-BLSTM \cite{Puigcerver2017}} & CTC\tnote{2} & $5.8$ \\
		& CTC\tnote{2} $\,$+ LM & $4.4$ \\[1mm]
		Fully-CNN \cite{YousefHS2018} & CTC\tnote{2} & $4.9$ \\
		\hhline{--||-}
		CNN-BLSTM + GRU \cite{PoulosValle2017} & Seq2Seq\tnote{3} & $16.9$ \\
		CNN-BLSTM + LSTM \cite{ChowdhuryVig2018} & Seq2Seq\tnote{3} & $8.1$ \\
		CNN-BLSTM + LSTM \cite{SueirasRSV2018} & Seq2Seq\tnote{4} & $8.8$ \\
		CNN-BGRU + GRU \cite{KangTRVFR2018} & Seq2Seq\tnote{4} & $6.9$ \\
		\hhline{--||-}
		Ours & Seq2Seq\tnote{2} & $4.87$ \\
		\hhline{--||-}
	\end{tabular}
	\begin{tablenotes}[para]
	\item [1] IAM-B partition
	\item [2] IAM-A partition
	\item [3] IAM-C partition\newline
	\item [4] IAM-A partition on word-level
	\end{tablenotes}
\end{threeparttable}
\vspace{-3mm}
\end{table}

\section{Conclusion}\label{sec:conclusion}
The paper successfully approaches the HTR task with an attention-based Seq2Seq architecture. The model combines a convolutional feature extractor with a recurrent neural network to encode both the visual information, as well as the temporal context in the input image, and uses a separate recurrent neural network to decode the actual character sequence. We make experimental comparisons between various attention mechanisms and positional encodings, in order to find an appropriate alignment between the input and output sequence. If the encoder conforms with the CTC framework, the model is able to retain an interpretable encoder output. This is achieved by training with a hybrid loss, which integrates the standard CTC loss as a local objective for the encoder, but is not necessary for the overall performance regarding HTR.

Overall, we obtain results which are competitive with the state-of-the-art on popular handwriting data sets, even without the use of a language model, and we beat any similar Seq2Seq approaches for HTR significantly.

The proposed model allows for potential future research about the direct integration of language resources or external memory modules. Augmenting the model with an external memory module that is expandable, without growing the number of trained parameters (see e.g. Labeled Memory Networks \cite{ShankarS2017}) could be beneficial for fast online model adaptation or the ability to cope with rare examples when training data is scarce. Furthermore, the Seq2Seq model could be trained in tandem with a pretrained language model (see e.g. Cold Fusion \cite{SriramJSC2017}) or one could try to pretrain the decoder as a stand-alone system on language resources and combine it with a suitable encoder afterwards. Nevertheless, whether a direct integration of an LM into the decoder will outperform CTC and a conventional LM remains to be seen.

\section*{Acknowledgment}
\ifthenelse{\blind}
{\textit{BLIND ACKNOWLEDGMENT}}
{This work was partially funded by the European Union's Horizon 2020 research and innovation programme under grant agreement No 674943 (READ -- Recognition and Enrichment of Archival Documents) and under grand agreement No 770299 (NewsEye).}


\bibliographystyle{IEEEtran}
%


\bibliography{lit}

\end{document}